\documentclass[10pt,twocolumn,letterpaper]{article}

\usepackage{iccv}
\usepackage{times}
\usepackage{epsfig}
\usepackage{graphicx}
\usepackage{amsmath}
\usepackage{amssymb}
\usepackage{subcaption}

\usepackage{authblk}

\usepackage[pagebackref=true,breaklinks=true,letterpaper=true,colorlinks,bookmarks=false]{hyperref}

\iccvfinalcopy 

\def\iccvPaperID{1837} 

\ificcvfinal\fi
\begin{document}

\title{Multilevel Context Representation for Improving Object Recognition}

\author[ ]{Andreas K\"olsch\textsuperscript{1}, Muhammad Zeshan Afzal\textsuperscript{1,2} and Marcus Liwicki\textsuperscript{1,3}\\\texttt {a\_koelsch12@cs.uni-kl.de, afzal@iupr.com, marcus.liwicki@unifr.ch}}
\affil[1]{MindGarage, University of Kaiserslautern}
\affil[2]{Insiders Technologies GmbH}
\affil[3]{University of Fribourg}

\renewcommand\Authands{ and }


\maketitle

\begin{abstract}

In this work, we propose the combined usage of low- and high-level blocks of convolutional neural networks (CNNs) for improving object recognition. While recent research focused on either propagating the context from all layers, e.g. ResNet, (including the very low-level layers) or having multiple loss layers (e.g. GoogLeNet), the importance of the features close to the higher layers is ignored. This paper postulates that the use of context closer to the high-level layers provides the scale and translation invariance and works better than using the top layer only. In particular, we extend AlexNet and GoogLeNet by additional connections in the top $n$ layers.
In order to demonstrate the effectiveness of the proposed approach, we evaluated it on the standard ImageNet task. The relative reduction of the classification error is around 1-2\% without affecting the computational cost.
Furthermore, we show that this approach is orthogonal to typical test data augmentation techniques, as recently introduced by Szegedy et al. (leading to a runtime reduction of 144 during test time).

\end{abstract}

\section{Introduction}
\label{sec:introduction}

\begin{figure}
\begin{subfigure}{.32\linewidth}
  \centering
    \includegraphics[width=0.9\linewidth]{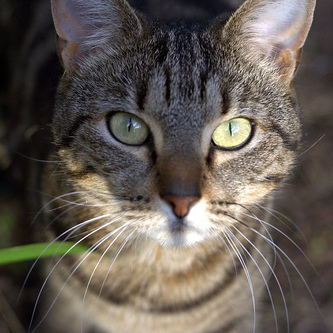}
    \captionsetup{justification=centering}
    \caption{``Tabby cat''}
\end{subfigure}
\begin{subfigure}{.32\linewidth}
  \centering
    \includegraphics[width=0.9\linewidth]{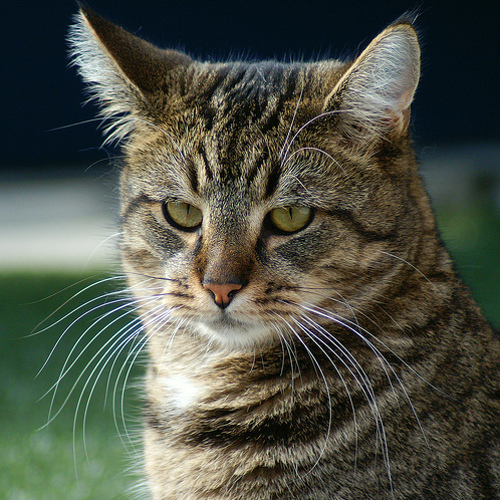}
    \captionsetup{justification=centering}
    \caption{``Tiger cat''}
\end{subfigure}
\begin{subfigure}{.32\linewidth}
  \centering
    \includegraphics[width=0.9\linewidth]{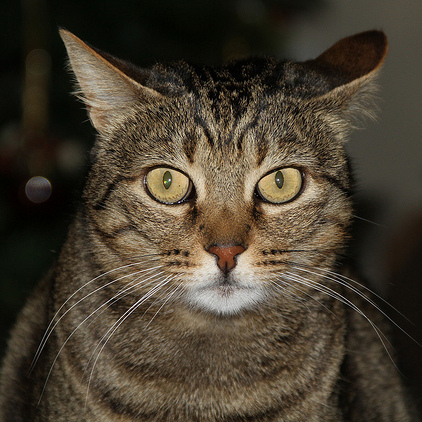}
    \captionsetup{justification=centering}
    \caption{``Egyptian cat''}
\end{subfigure}
\caption{Three similar images from different classes}
\label{fig:different_class}
\end{figure}
\begin{figure}
\begin{subfigure}{.32\linewidth}
  \centering
    \includegraphics[width=0.9\linewidth]{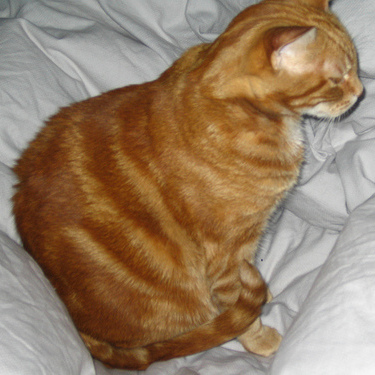}
    \captionsetup{justification=centering}
\end{subfigure}
\begin{subfigure}{.32\linewidth}
  \centering
    \includegraphics[width=0.9\linewidth]{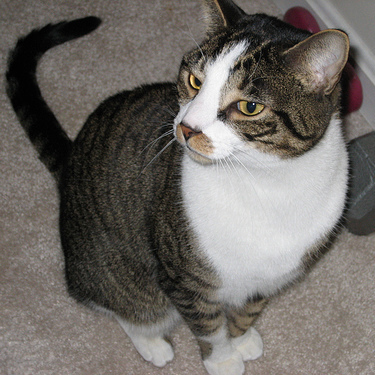}
    \captionsetup{justification=centering}
\end{subfigure}
\begin{subfigure}{.32\linewidth}
  \centering
    \includegraphics[width=0.9\linewidth]{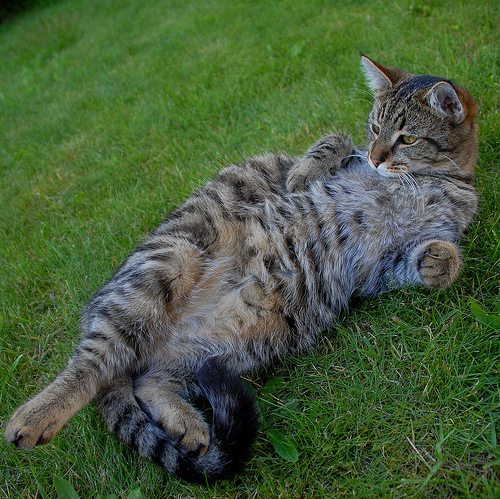}
    \captionsetup{justification=centering}
\end{subfigure}
\caption{Even though they look quite different, all three of these images belong to the class ``Tiger cat''}
\label{fig:same_class}
\end{figure}

While it is quite easy for humans to distinguish between objects in an image, it can be a very challenging task for a computer. Not only can objects appear in different sizes and angles, but also backgrounds and lighting conditions may vary and many other factors can change the way an object is displayed~\cite{zeiler2013hierarchical}.
Furthermore, some object classes have a very low inter-class variance (cf.\ Fig.~\ref{fig:different_class}), while other classes might have a very high intra-class variance (cf.\ Fig.~\ref{fig:same_class}) \cite{he2015delving, goring2013fine}. Consequently, both the details of an image and the greater context are important for successful classification.
Another factor, that makes image classification hard is the presence of several objects in one image (cf.\ Fig.~\ref{fig:multiple_objects}).
When it comes to realistic applications of object recognition models, the task is exacerbated even further by computational constraints. Some applications, e.g. in autonomous cars, require very fast, yet accurate predictions~\cite{teichman2011practical, lecun2010convolutional}, while other applications, especially those in mobile devices, constrain the amount of available memory and processing power.

\begin{figure}
\centering
\includegraphics[width=\linewidth]{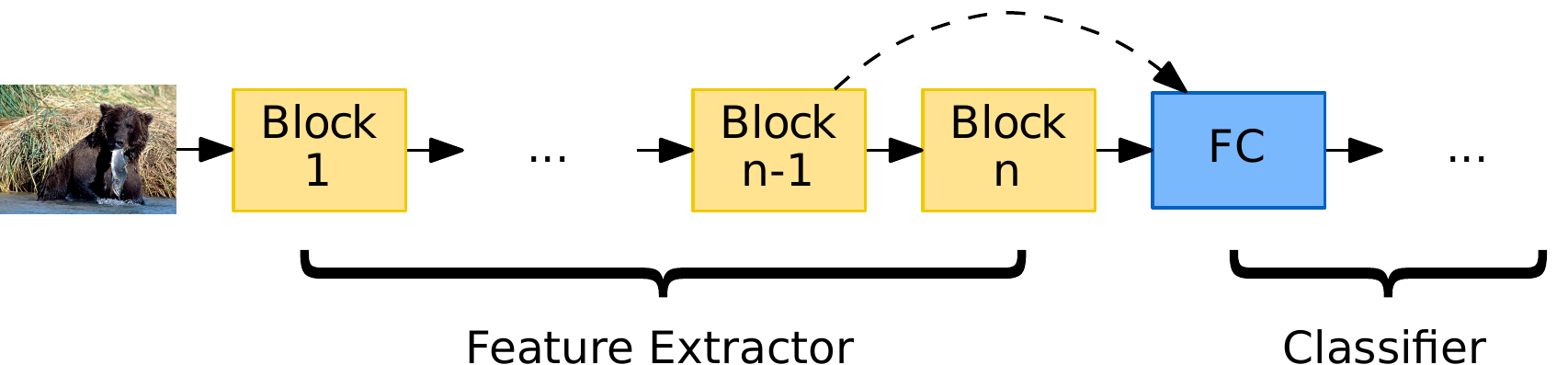}
\caption{Contribution of this work: Instead of just using the last CNN layer (Block $n$), we use connections to more high-level layers as well in order to include various representations in the recognition.}
\label{fig:contribution}
\end{figure}

In the last years, neural networks have had great success in many tasks related to computer vision. For the task of object recognition, particularly convolutional neural networks (CNNs) have become the state of the art~\cite{gu2015recent}. Thanks to now available computational resources and efficient GPU implementations~\cite{jia2014caffe, collobert2011torch7}, it is also possible to train very deep CNNs.
In 2012, Krizhevsky et al. proposed an eight-layer deep CNN~\cite{alexnet}, which, due to its groundbreaking performance, laid the foundation of using these networks for image classification. Two years later, 19-layer~\cite{vgg} and 22-layer networks~\cite{googlenet} were presented, which performed even better. In 2015, He et al. proposed an even deeper network architecture with 152 layers~\cite{resnet}, and most recently the development of ever deeper CNNs culminated with PolyNet with several hundred layers~\cite{zhang2016polynet} (see Section~\ref{sec:related}).

While this trend towards deeper networks is evident, the basic architecture of CNNs has never changed fundamentally. They always consist of a number of stacked convolutional layers with a non-linear squashing function~\cite{hinton2012improving, lecun1995convolutional, dahl2013improving}, optional local response normalization and pooling for feature extraction~\cite{boureau2010theoretical, ciresan2011flexible, scherer2010evaluation}, followed by a number of fully-connected layers for classification.
As illustrated by Zeiler and Fergus, the features extracted by the convolutional layers in the network correspond to more complex patterns from layer to layer~\cite{Zeiler2014}. While the features of the lower level layers correspond to dots or edges, the higher level features correspond to
patterns, such as faces or entire objects. In the case of AlexNet~\cite{alexnet}, Zeiler and Fergus describe the features extracted by the two highest level convolutional layers of the network, as highly ``class-specific''~\cite{Zeiler2014}.

However, to the best of the authors' knowledge, it has not yet been tried to make the information from multiple high-level layers directly available to the classification part of the network to test, whether or not the additional features provide complementary information.
This is the main idea of this paper.
As such, we hope that providing the classification part of the network with more data leads to better classification results while keeping the additional computational costs for both training and prediction minimal.

The contributions of this work are three-fold.
Firstly, we perform some initial experiments motivating the idea of adding connections from multiple levels.
Secondly, we present two novel CNN architectures that explicitly use the features extracted by multiple convolutional layers, for classification. These networks do not rely on pre-training, but provide standalone solutions that can be trained from scratch.
Thirdly, we evaluate both the performance and computational costs of the novel architectures and compare them to existing approaches.

The remaining sections are organized as follows. Section~\ref{sec:related} gives an overview of related work. Section~\ref{sec:poc} shows the validity of the approach in a constrained setup. In section~\ref{sec:experiments} the novel network architectures and the training methodology are described in detail and large scale experiments are performed. The results are presented in section~\ref{sec:results} which is followed by the discussion in section~\ref{sec:discussion} and conclusion in section~\ref{sec:conclusion}.

\section{Related Work}
\label{sec:related}

\definecolor{red}{RGB}{255, 0, 0}
\definecolor{color1}{rgb}{0,0.5,0}
\definecolor{color2}{rgb}{0,0,1}
\definecolor{color3}{rgb}{1,0,0}
\definecolor{color4}{rgb}{1,0,1}
\definecolor{color5}{rgb}{1,1,0}
\definecolor{color6}{rgb}{0,1,1}

\begin{figure}
\centering
\includegraphics[width=\linewidth]{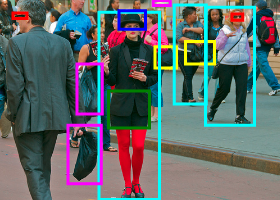}
\caption{An image showing objects of different classes: \textcolor{color1}{miniskirt}, \textcolor{color2}{Hat with a wide brim}, \textcolor{color3}{sunglasses}, \textcolor{color4}{plastic bag}, \textcolor{color5}{purse}, \textcolor{color6}{person}}
\label{fig:multiple_objects}
\end{figure}

Both Convolutional Neural Networks (CNNs) and Recurrent Neural Networks (RNNs) have been used for solving key computer vision problems~\cite{lecun2015deep}. Some example areas such as object recognition and detection~\cite{alexnet, googlenet, resnet, zhang2016polynet, visin2015renet}, semantic segmentation ~\cite{chen2014semantic, lin2016efficient, long2015fully, noh2015learning}, describing images with text~\cite{karpathy2015deep, xu2015show, vedantam2017context, donahue2015long} and also generating the images using text~\cite{goodfellow2014generative, mirza2014conditional, odena2016conditional, reed2016generative, zhang2016stackgan} show the effectiveness of the deep learning models for solving challenging tasks. The RNNs did not receive much attention for object recognition and the work in this area is rather sparse due to the following reasons: First by the virtue of their architecture that is suitable for temporal data processing. Secondly, due to the temporal dependencies, the processing of RNNs is inefficient in comparison to CNNs. However, the work of Visin et al.~\cite{visin2015renet} had signified that contextual processing of pixels can help to recognize the objects better. Their proposed architecture was named as ReNet and it was recommended as an alternative approach to CNNs. In another work, Biswas et al.~\cite{biswas2014using} concatenated different variations (degradations) of the same object and processed it temporally. They used each variation of the object as one time step for RNNs. However, their experiments are limited to MNIST~\cite{lecun-mnisthandwrittendigit-2010} and COIL-20~\cite{nene1996columbia} data sets and the validity and the computational viability of their approach remains undiscovered for large data sets e.g., ImageNet~\cite{ILSVRC15}. A combination of CNNs and RNNs has also been reported by Linag et al.~\cite{liang2015recurrent}. Their proposed architecture relies both on the efficiency of CNNS and power of contextual processing of RNNs. As our proposed work aims to extend existing deep CNN architectures, the following paragraphs briefly describe the state-of-the-art deep CNN architectures for object recognition.

One well-known benchmark for several visual tasks, including image classification, is the annual ImageNet Large Scale Visual Recognition Challenge (ILSRVC)~\cite{ILSVRC15}. It allows the participating teams to compare the performance of their developed models. The data set used for the competition consists of 1.000 image classes containing 1.2 million labeled images for training, 50.000 labeled validation images and 100.000 unlabeled images which are used to compare the performance of the submitted entries.

In 2012, Krizhevsky et al.\ have, for the first time, used a deep CNN in the ILSVRC and won the image classification with a significant margin, outperforming all traditional methods~\cite{alexnet}. The winning architecture consists of five convolutional layers which are followed by three fully connected layers. The first and second layer use local response normalizations and max-pooling before passing the activations to the next layer. The fifth convolutional layer is again followed by a max-pooling layer which provides the input to the fully-connected layers. To prevent overfitting, the fully-connected layers use dropout~\cite{hinton2012improving}. Rectified linear units (ReLU) are used as activation function in all layers \cite{nair2010rectified}. AlexNet achieves a top-5 error rate of $16.422\%$ on the ILSVRC test data set.

\begin{figure}
\begin{center}
\centerline{\includegraphics[width=\columnwidth]{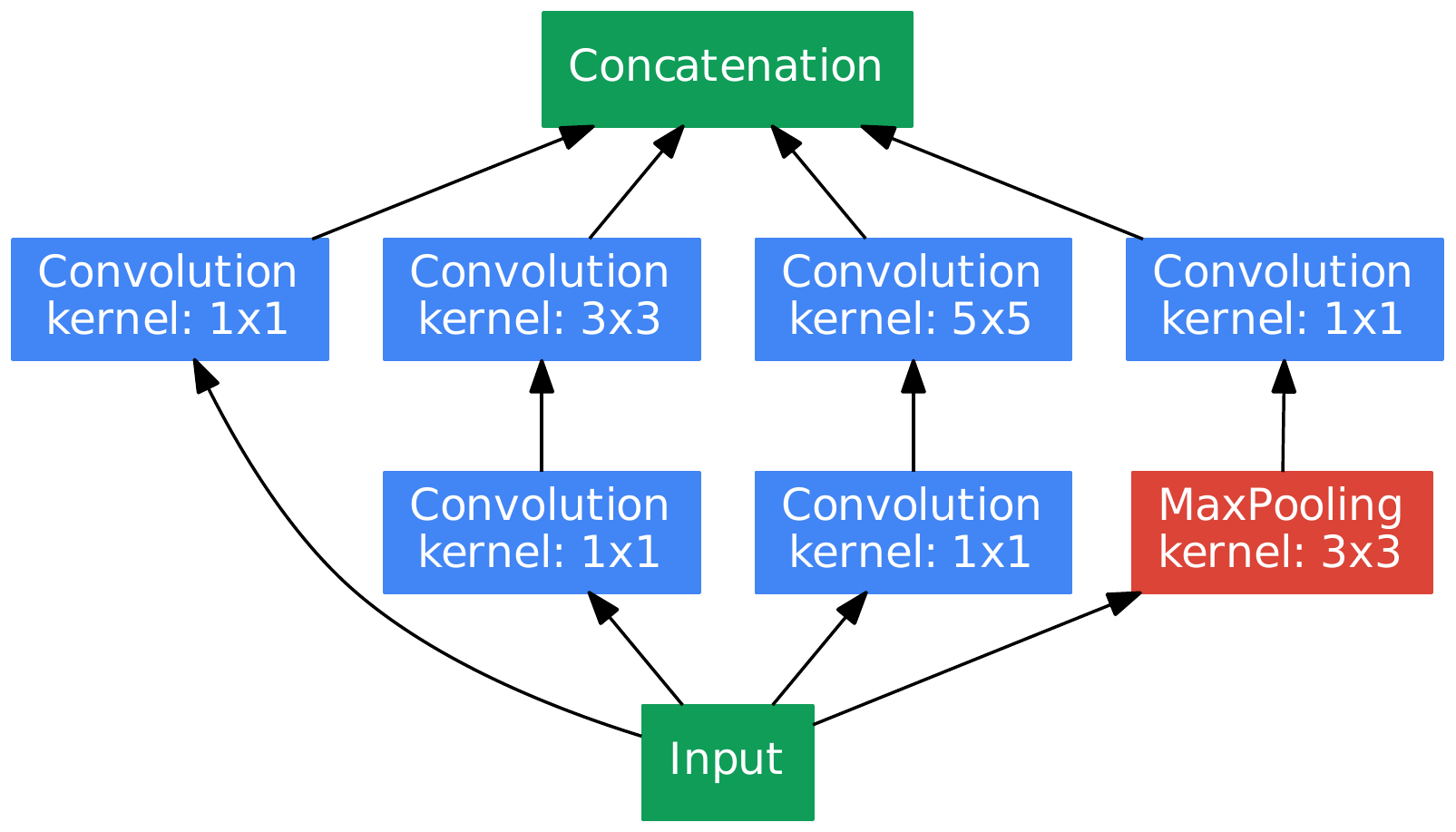}}
\caption{Building block of the Inception architecture~\cite{googlenet}.}
\label{fig:bb}
\end{center}
\vskip -0.2in
\end{figure} 

Since then, many other teams have developed a variety of different CNN architectures to further reduce the classification error of these networks~\cite{resnet, vgg, googlenet}. First of all, Szegedy et al. presented the 22-layer GoogLeNet in 2014, which achieved a top-5 error rate of $6.656\%$ on the ILSVRC 2014 test data set with an ensemble of 7 networks. The network is using the Inception architecture~\cite{googlenet} which is different from other approaches, as it employs not stacked convolutional layers, but stacked building blocks which themselves consist of multiple convolutional layers (cf.\ Fig.~\ref{fig:bb}). Therefore, it is a network-in-network approach~\cite{lin2013network}. As the network is so deep, it employs two auxiliary loss layers during training to allow for efficient backpropagation of the error.

\begin{figure}
\begin{center}
\centerline{\includegraphics[width=0.6\columnwidth]{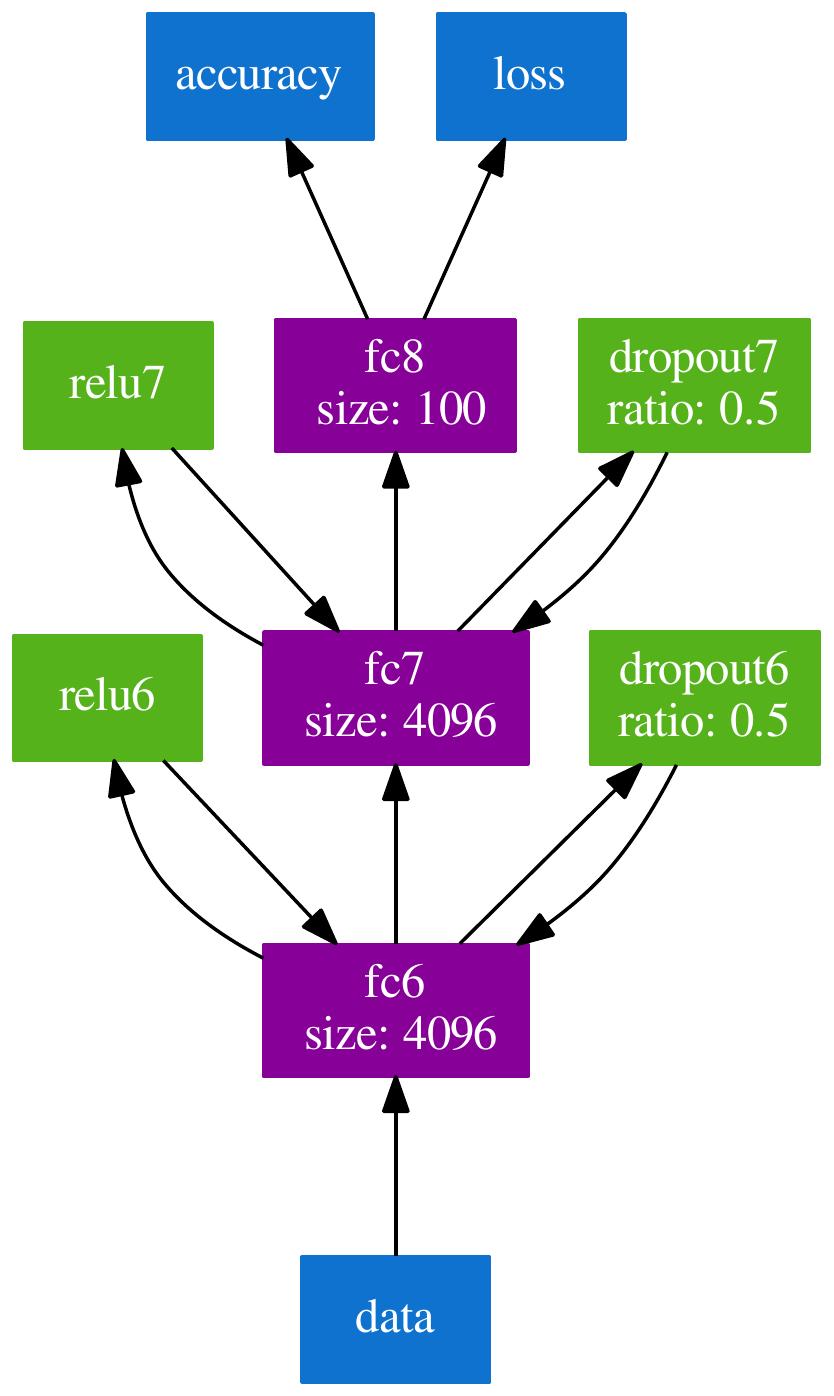}}
\caption{Network architecture used for the experiments presented in section~\ref{sec:poc}. The data layer holds the features extracted beforehand. The fc8 layer is adjusted in size to match the number of classes used in the experiments.}
\label{fig:experiment}
\end{center}
\vskip -0.2in
\end{figure}

Note that the general idea of adding skip-connections (so-called residual connections) has been recently introduced by He et al. in~\cite{resnet, he2016identity}.
However, the motivation and effects are fundamentally different.
In ResNet, the skip-connections go through all the layers, letting the layers mainly pass the information through the network with minor additive terms.
This mainly overcomes the vanishing gradient problem and allows learning architectures with hundreds of layers.
Contrary, the main idea in this paper is to use the extracted features of multiple layers directly for classification.
The only work which comes close to our idea is the proposal of Center-Multilayer Features (CMF) as proposed by Seuret et al.\cite{seuret2016n}.
They used stacked convolutional auto-encoders (SCAE) for classifying an image or the center pixel of a patch. This idea is similar to our pre-study (section~\ref{sec:poc}) where we use multiple feature representations from fine-tuned networks. However, their network architecture is shallow and they use features from all the layers.

Apart from novel network architectures, a lot of work has been done to understand how the networks are learning and how they can be improved~\cite{he2015convolutional, mishkin2016systematic}. Especially visualization techniques have helped to get an understanding of the convolutional layers~\cite{Zeiler2014}.
Despite all the research that has been done to develop new CNN architectures, novel architectures typically employ more layers or make use ensembling techniques \cite{zhou2002ensembling}. However, both approaches typically require more computational resources at training and inference time.

\section{Pre-study}
\label{sec:poc}
To investigate, whether it is worth pursuing the idea of using features from multiple layers, we run a set of small-scale experiments. We use a publicly available model\footnote{\url{https://github.com/BVLC/caffe/tree/master/models/bvlc_alexnet}} of AlexNet that is pre-trained on ImageNet. This model is used to extract and store the activations generated by the fourth and fifth convolutional layers (conv4 and conv5) during a forward pass of the images. Then, we train a three-layer fully-connected neural network on these activations which, except for the last layer, equals the fully-connected part of AlexNet (cf.\ Fig.~\ref{fig:experiment}). As is common, the last layer is adjusted in size to match the number of classes. We repeat the experiment with several subsets of the ILSVRC data set. The network is trained and evaluated on the activations of 10, 20, 50 and 100 randomly selected classes of the data set.

\begin{table}
\caption{Accuracy achieved by networks on validation sets after training with inputs from conv5 only and inputs from conv4 (normalized) and conv5.}
\vskip 0.15in
\begin{center}
\begin{small}
\begin{tabular}{ l | c | c } 
& conv5 & conv4 \& conv5 \\
\hline
10 classes & $90.60\%$ & $91.60\%$\\
\hline
20 classes & $83.40\%$ & $84.10\%$\\
\hline
50 classes & $78.72\%$ & $80.12\%$\\
\hline
100 classes & $68.36\%$ & $71.10\%$\\
\end{tabular}
\label{tab:experiment}
\end{small}
\end{center}
\end{table}

To get a baseline, we run a set of experiments, in which the network is trained solely on the activations from conv5 that were extracted before. In another set of experiments, we use the same classes, but train the network on the activations of conv4 concatenated with conv5. It turns out, we have to L2-normalize the activations of conv4 to get useful results. With unmodified values, the network fails to learn.

Table~\ref{tab:experiment} shows, that the networks trained on features from conv4 and conv5 yielded a better performance on the validation set than the networks which were trained on the features from conv5 only.

\section{Large Scale Experiments}
\label{sec:experiments}

With the results from section~\ref{sec:poc} we step the experiments up and train entire CNNs from scratch on the ILSVRC data set.

\subsection{Network Architectures}

\begin{figure}
\begin{center}
\centerline{\includegraphics[height=0.95\textheight]{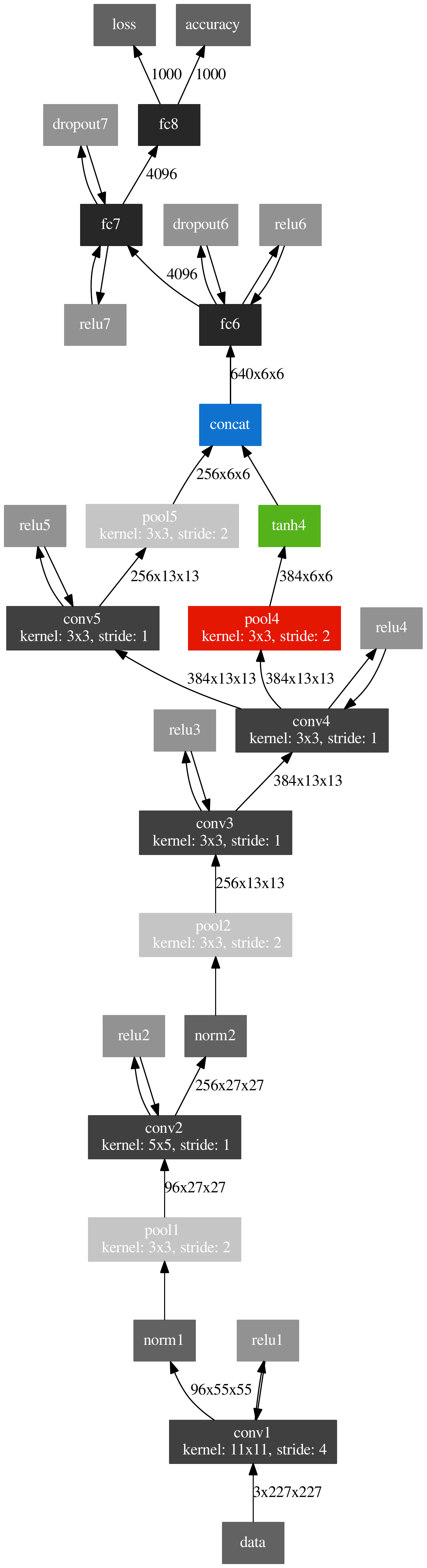}}
\caption{AlexNet++ architecture with unchanged AlexNet part in gray and the new part colored.}
\label{fig:alex++}
\end{center}
\end{figure}

We propose two neural network architectures in this section.
The network architectures are strict extensions of the existing networks AlexNet and GoogLeNet. Unlike most CNNs, including AlexNet and GoogLeNet, the proposed networks feed the classification part of the network with information not only from the highest-level convolutional layer, but with information from the two highest-level convolutional layers. We call the enhanced versions of these networks AlexNet++ and GoogLeNet++.

\subsubsection{AlexNet++}
The original AlexNet is an eight-layer deep CNN in which each layer processes the features extracted by the subjacent layer only. As already said, the proposed AlexNet++ differs from the original AlexNet in the way, that the first fully connected layer (fc6) is not only presented with the activations from conv5, but also with activations from conv4 (cf.\ Fig.~\ref{fig:alex++}). Since after max-pooling, the conv5 layer has only $256\times6\times6$ values, but conv4 has $384\times13\times13$ values, we have to balance the number of activations passed to the fc6 layer. Therefore, we add a max-pooling of conv4, before concatenating the values with the conv5 values. Furthermore, to resemble the normalization that was done in the small scale experiments (cf.\ Section~\ref{sec:poc}), we use \emph{tanh} after the pooling to compress the activations. A detailed comparison of the layer sizes of AlexNet and AlexNet++ is given in table~\ref{tab:alexnet_change}.

\begin{table}
\centering
\caption{Input size of the layers of AlexNet and AlexNet++}
\label{tab:alexnet_change}
\begin{tabular}{r|c|c}
 & AlexNet & AlexNet++ \\ \hline
conv1 & $3\times227\times227$ & $3\times227\times227$ \\ \hline
pool1 & $96\times55\times55$ & $96\times55\times55$ \\ \hline
conv2 & $96\times27\times27$ & $96\times27\times27$ \\ \hline
pool2 & $256\times27\times27$ & $256\times27\times27$ \\ \hline
conv3 & $256\times13\times13$ & $256\times13\times13$ \\ \hline
conv4 & $384\times13\times13$ & $384\times13\times13$ \\ \hline
conv5 & $384\times13\times13$ & $384\times13\times13$ \\ \hline
pool5 & $256\times13\times13$ & $256\times13\times13$ \\ \hline
fc6 & $\textcolor{red}{256\times6\times6}$ & $\textcolor{red}{640\times6\times6}$ \\ \hline
fc7 & $4096$ & $4096$ \\ \hline
fc8 & $4096$ & $4096$ \\ \hline
prob & $1000$ & $1000$
\end{tabular}
\end{table}

\subsubsection{GoogLeNet++}
GoogLeNet is the 22-layer network architecture, that won the ILSVRC2014 \cite{googlenet}. Unlike AlexNet, the architecture consists of stacked building blocks which consist of convolutional layers (cf.\ Fig.~\ref{fig:bb}).
Due to its depth, the original architecture employs three softmax classifiers for efficient error backpropagation during training. At inference time, only the main classifier is used. This classifier gets its input from one fully connected layer, which in turn processes the input from the highest level building block.

\begin{figure}
\begin{center}
\centerline{\includegraphics[width=\columnwidth]{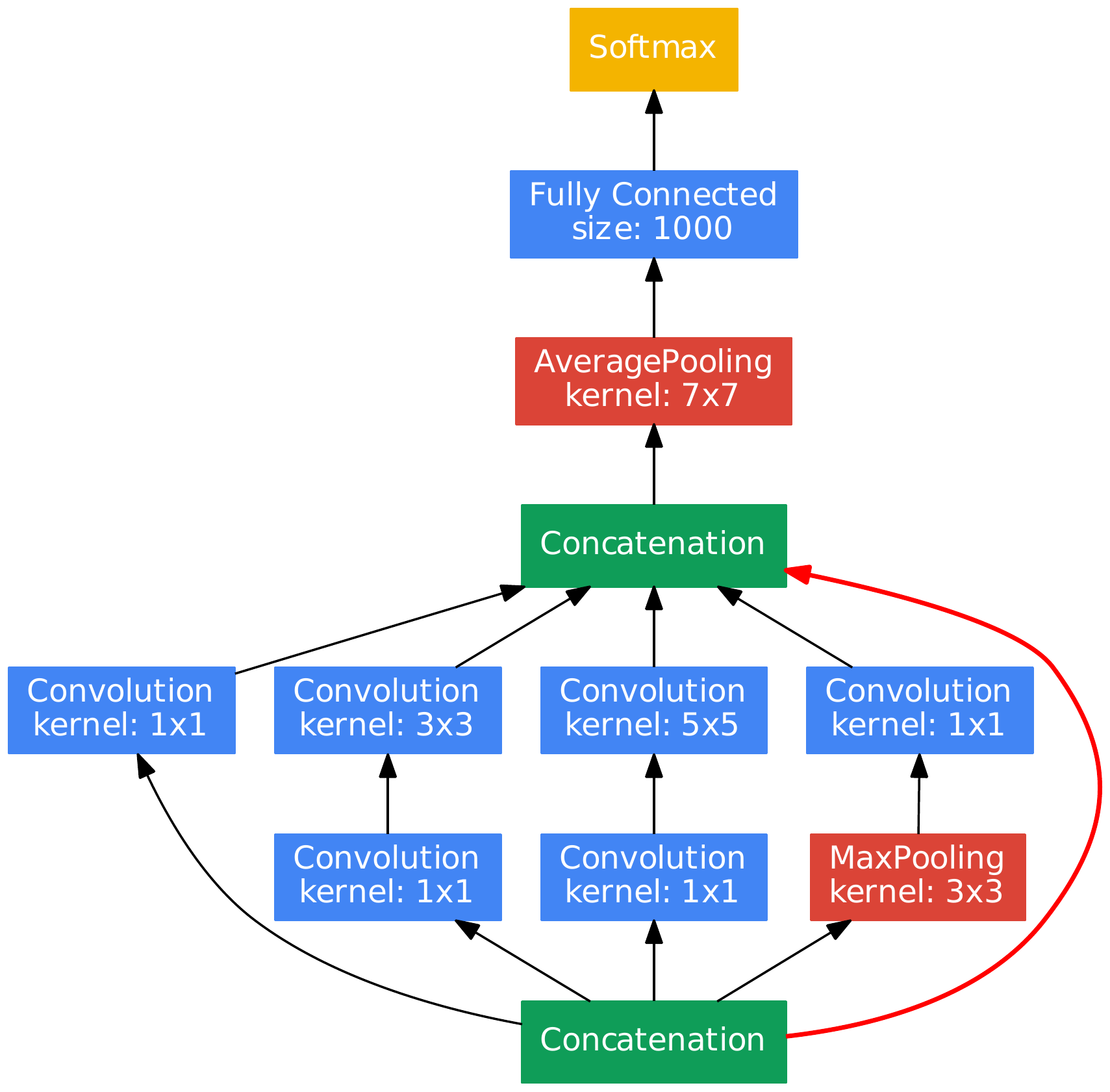}}
\caption{Top part of the GoogLeNet++ architecture. The new connection is marked red. The lower layers of the network are built as proposed by Szegedy et al.~\cite{googlenet}.}
\label{fig:googlenet++}
\end{center}
\vskip -0.2in
\end{figure}

The proposed GoogLeNet++ architecture concatenates the activations of the two highest level building blocks, before they are passed to the fully connected layer (cf.\ Fig.~\ref{fig:googlenet++}). Table~\ref{tab:googlenet_change} gives a detailed comparison of the layer sizes of GoogLeNet and GoogLeNet++.

\begin{table}
\centering
\caption{Input size of the layers of GoogLeNet and GoogLeNet++}
\label{tab:googlenet_change}
\begin{tabular}{r|c|c}
 & GoogLeNet & GoogLeNet++ \\ \hline
conv1 & $3\times224\times224$ & $3\times224\times224$ \\ \hline
pool1 & $64\times112\times112$ & $64\times112\times112$ \\ \hline
conv2 & $64\times56\times56$ & $64\times56\times56$ \\ \hline
pool2 & $192\times56\times56$ & $192\times56\times56$ \\ \hline
inception3a & $192\times28\times28$ & $192\times28\times28$ \\ \hline
inception3b & $256\times28\times28$ & $256\times28\times28$ \\ \hline
pool3 & $480\times28\times28$ & $480\times28\times28$ \\ \hline
inception4a & $480\times14\times14$ & $480\times14\times14$ \\ \hline
inception4b & $512\times14\times14$ & $512\times14\times14$ \\ \hline
inception4c & $512\times14\times14$ & $512\times14\times14$ \\ \hline
inception4d & $512\times14\times14$ & $512\times14\times14$ \\ \hline
inception4e & $528\times14\times14$ & $528\times14\times14$ \\ \hline
pool4 & $832\times14\times14$ & $832\times14\times14$ \\ \hline
inception5a & $832\times7\times7$ & $832\times7\times7$ \\ \hline
inception5b & $832\times7\times7$ & $832\times7\times7$ \\ \hline
pool5 & $\textcolor{red}{1024\times7\times7}$ & $\textcolor{red}{1856\times7\times7}$ \\ \hline
fc & $\textcolor{red}{1024\times1\times1}$ & $\textcolor{red}{1856\times1\times1}$ \\ \hline
prob & $1000$ & $1000$
\end{tabular}
\end{table}

\subsection{Training Details}
We train a version of AlexNet++ and GoogLeNet++ from scratch on the ILSVRC training data. To allow for a fair comparison, the unchanged versions of AlexNet and GoogLeNet are also trained on the same training data and setup. All networks are trained on a single NVidia Tesla K20X using the Caffe framework~\cite{jia2014caffe}.

AlexNet and AlexNet++ are trained for 90 epochs using stochastic gradient descent with an initial learning rate of $0.01$ which is divided by 10 after 30 and 60 epochs. The momentum is set to $0.9$ and the weight decay is $0.0005$ as proposed by Krizhevsky et al.~\cite{alexnet}. Both networks use a batch size of 256 images to speed up the training. As proposed in the original architecture, we initialize the weights with a gaussian distribution with a mean of zero and a standard deviation of $0.01$. The biases are initialized differently from the original approach. While Krizhevsky et al.\ set the biases of the second, fourth, and fifth convolutional layers and the biases of the fully-connected layers to $1$, we set them to $0.1$, as the network fails to learn with biases set to~1\footnote{\url{https://github.com/BVLC/caffe/tree/master/models/bvlc_alexnet}}.

The GoogLeNet and GoogLeNet++ networks are trained for five million iterations with a batch size of 32, i.e. for 133 epochs. We train the networks with stochastic gradient descent with a polynomial update of the learning rate, as proposed by Sergio Guadarrama\footnote{\url{https://github.com/BVLC/caffe/tree/master/models/bvlc_googlenet}}. This means, the learning rate is updated at every iteration to 
\begin{equation}
    lr = initial\_lr * \left( 1 - \frac{iter}{max\_iter}\right) ^ {power}
\end{equation}
In our case, the initial learning rate is set to $0.01$ and power is $0.5$. The momentum is $0.9$, weight decay is set to $0.0002$. To initialize the network, we use \emph{normalized initialization}~\cite{glorot2010understanding}.

\subsubsection{Data Augmentation}
\begin{figure}
\begin{subfigure}{.49\linewidth}
  \centering
    \includegraphics[width=0.9\linewidth]{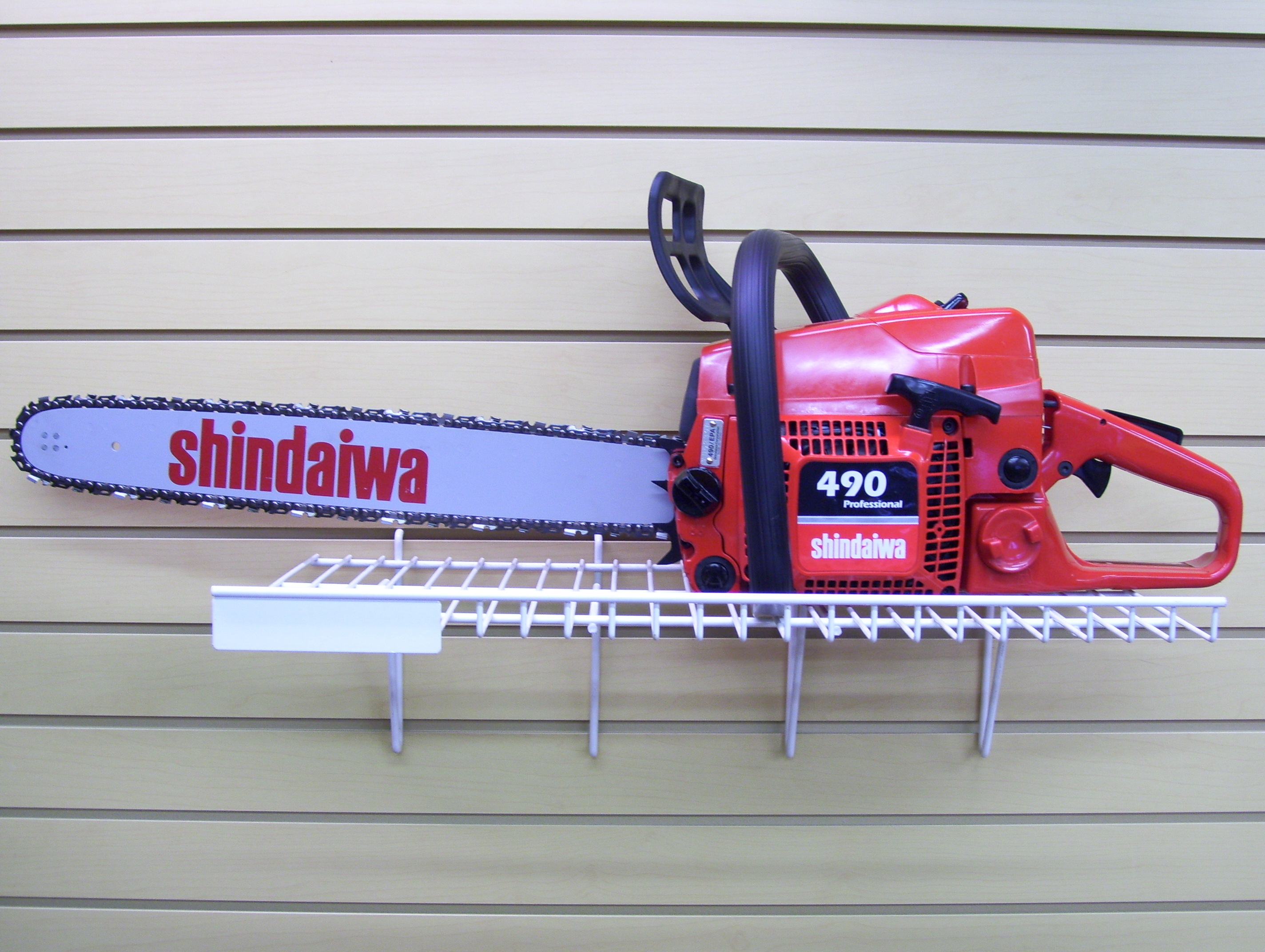}
    \captionsetup{justification=centering}
    \caption{ILSVRC image ($2848\times2144$~pixels)}
\end{subfigure}
\begin{subfigure}{.49\linewidth}
  \centering
    \includegraphics[width=0.9\linewidth]{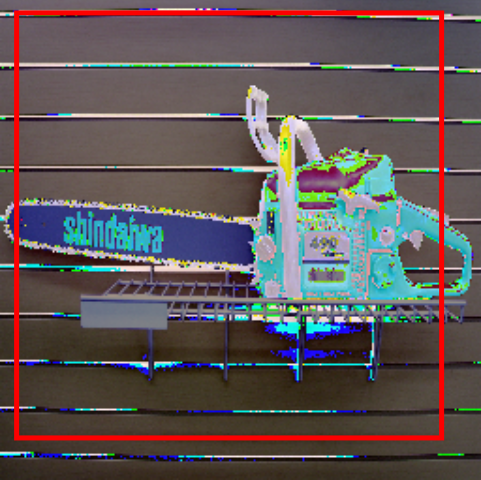}
    \captionsetup{justification=centering}
    \caption{Preprocessed image ($256\times256$~pixels)}
\end{subfigure}
\caption{Images are scaled to $256\times256$ pixels, the mean pixel value is subtracted and the image is randomly cropped to the network's input size.}
\label{fig:data_augmentation}
\end{figure}

The four networks described above are trained on the ILSVRC training data set with no additional data. To artificially enlarge the data set, we use a rather simple data augmentation technique. Namely, we resize all images to $256\times256$ pixels, subtract the mean pixel value in a preprocessing step and then randomly crop patches in the size of the network input from these, i.e. $227\times227$ pixels for AlexNet, $224\times224$ pixels for GoogLeNet. Finally, we randomly mirror the images horizontally. An example is given in Fig.~\ref{fig:data_augmentation}.

Note, that Krizhevsky et al.\ and Szegedy et al.\ use more aggressive data augmentation techniques. However, in the case of GoogLeNet, the given details are very vague and not described in a reproducible way. In the case of the original AlexNet approach, the rectangular images are preprocessed by scaling them such that the shorter edge measures $256$ pixels and then a $256\times256$ patch is cropped from the center. This scaling preserves the aspect ratio, but does not use the outer regions of an image at all.
We decided to not perform more aggressive data augmentation techniques as our idea is orthogonal to the augmentation ideas, i.e., even if more data augmentation is performed, it is still possible to use the Multilevel Context representation.

\section{Benchmarking and Results}
\label{sec:results}

To compare our extended versions of AlexNet and GoogLeNet with the original ones, we use the labeled ILSVRC validation data set. Remember, we not only set out to improve the classification performance of the existing networks, but also, to keep the additional computational costs low. Therefore, this section covers both, the classification accuracy of the models and the time needed for classification and training.

\begin{table*}[t]
\caption{Accuracy achieved by networks on validation sets using center crops only and averaging over 144 crops as described by Szegedy et al. \cite{googlenet}}
\vskip 0.15in
\begin{center}
\begin{small}
\begin{tabular}{ l || c | c || c | c} 
& \multicolumn{2}{c||}{Top-1} & \multicolumn{2}{c}{Top-5} \\
\cline{2-5}
& center crop & 144 crops & center crop & 144 crops \\
\hline
AlexNet & $58.17\%$ & $58.54\%$ & $80.82\%$ & $81.18\%$ \\
\hline
AlexNet++ & $58.47\%$ & $58.93\%$ & $81.25\%$ & $81.47\%$ \\
\hline
GoogLeNet & $68.89\%$ & $68.86\%$ & $88.97\%$ & $89.26\%$ \\
\hline
GoogLeNet++ & $69.42\%$ & $69.05\%$ & $89.25\%$ & $89.35\%$\\
\end{tabular}
\label{tab:accuracy}
\end{small}
\end{center}
\end{table*}

\begin{table*}[t]
\caption{Time needed for passing one batch through the network. The batch size for AlexNet and AlexNet++ is 256, the batch size for GoogLeNet and GoogLeNet++ is 32.}
\vskip 0.15in
\begin{center}
\begin{small}
\begin{tabular}{ l | c | c | c} 
& Forward & Backward & $\Sigma$ \\
\hline
AlexNet & $254.87$ ms & $654.04$ ms & $909.01$ ms \\
\hline
AlexNet++ & $271.29$ ms & $686.76$ ms & $958.16$ ms \\
\hline
GoogLeNet & $112.26$ ms & $283.31$ ms & $395.71$ ms \\
\hline
GoogLeNet++ & $112.51$ ms & $284.76$ ms & $397.42$ ms \\
\end{tabular}
\label{tab:benchmark}
\end{small}
\end{center}
\end{table*}

We use two evaluation methods to measure the accuracy of the networks. First, we classify the validation images using a single-crop technique, i.e., we resize them to $256\times256$ pixels and use the center crop. Secondly, the networks are tested with the multi-crop technique proposed in \cite{googlenet}. Specifically, the images are resized such that the shorter dimension is 256, 288, 320 and 352 pixels long. Then 3 square crops are taken from each of these images, i.e., the left, center and right part for landscape pictures or the top, center and bottom part for portrait pictures. For each of these square images the 4 corner crops, the center crop and a resized image with the shape of the network input is produced. Finally, all of these images are mirrored as well. In sum, this multi-crop technique produces $4\times3\times6\times2 = 144$ crops. To evaluate the network performance with this approach, we average over the probability vectors of the 144 predictions.

Table~\ref{tab:accuracy} presents the accuracy on the validation data set which is achieved by the four trained networks. As can be seen, the extended networks outperform the original versions with all evaluation techniques. Furthermore, the accuracy achieved by the new models with only one crop is at least comparable to the accuracy achieved by the original models using 144 crops. In the case of GoogLeNet, the Top-1 accuracy of the new network using only one crop is even more than $0.5\%$ higher than the accuracy of the original network using 144 crops (this corresponds to around $2\%$ relative error reduction).

Table~\ref{tab:benchmark} shows the time needed for forward and backward passes through the different networks. Obviously, the timings are specific to hardware and software. In our case, all timings are made using caffe 1.0rc3 and a NVidia Tesla K20X. 
AlexNet++ takes about 5\% longer for training, i.e., combined forward and backward pass. However, the accuracy achieved by AlexNet++ using just center crops is comparable and in the case of top-5 accuracy even higher than what AlexNet achieves using 144 crops (see Table~\ref{tab:accuracy}). Given this result, it is actually reasonable to say AlexNet++ is more than 100 times faster to reach a comparable accuracy.

For GoogLeNet++, the difference in runtime for both forward and backward pass is negligibly small. This is due to the fact, that the GoogLeNet architecture already has a concatenation layer before the fully connected layer and thus, no new layers have to be added to combine the two highest level building blocks (cf.\ Fig.~\ref{fig:googlenet++}). A combined forward- backward pass with GoogLeNet++ is only $0.4$\% slower than with the original GoogLeNet. Therefore, training the extended network takes only marginally longer. At inference time, the difference decreases to a mere $0.2$\% which makes the GoogLeNet++ very well-suited for productive usage.

\section{Discussion}
\label{sec:discussion}

Our approach postulates that the higher level layers are useful for achieving better performance both in terms of recognition accuracy and computational cost. In order to demonstrate the validity of the claim, we performed a simple experiment using AlexNet. In this experiment, the features from all of the convolutional layers were concatenated. Two observations were made: first, the training took significantly longer (almost $3$ times) and secondly, the accuracy was suboptimal (almost $2$\% less in terms of absolute error). The best performance was achieved by concatenating the features of the top $2$ layers.

The networks trained in this work could not reproduce the drastic error reduction that Szegedy et al. have reported \cite{googlenet} by using the $144$ image crops at test time.
The main reason for this is that we have used less aggressive data augmentation during training or due to different weight initializations. Unfortunately, Szegedy et al.\ do not report the details on how the GoogLeNet model was initialized.
Also, they report error reduction by ensembling multiple trained networks.
However, this is not in the scope of this paper. The main purpose of this paper is to show that features from lower level convolutional layers that are close to the highest level provide useful and complementary information for classification.

Noteworthy, as shown in this paper, when training our networks with the same data augmentation and hyper-parameters, the extended architectures presented in this work outperform the original ones.
This applies to both networks (AlexNet and GoogLeNet) in both evaluation scenarios (no data augmentation during testing or using $144$ test samples).

\section{Conclusion}
\label{sec:conclusion}

In this paper, we presented a successful approach for extending existing CNN architectures in order to boost their classification performance. We have demonstrated the effectiveness of this approach by enhancing the network architectures of AlexNet and GoogLeNet and training them from scratch on the ILSVRC data set. We consider networks that are totally different in nature to prove the generality of the proposed approach. Also, it is shown that at almost no additional cost, the relative error rates of the original networks decrease by up to 2\%.
This fact makes the extended networks a very well suited choice for usage in production environments.
The quantitative evaluation signifies that the new approach could be, at inference time, $144$ times more efficient than the current approaches while maintaining comparable performance.
The proposed approach is not limited to any one of the architectures. We plan to extend the experiments for recurrent and convolutional-recurrent neural networks.

{\small
\bibliographystyle{ieee}
\bibliography{literature}
}

\end{document}